\documentclass[10pt,twocolumn,letterpaper]{article}

\usepackage{cvpr}              % To produce the CAMERA-READY version
\usepackage{graphicx}
\usepackage{amsmath}
\usepackage{amssymb}
\usepackage{booktabs}
\usepackage{amsmath}
\usepackage{amssymb}
\usepackage{mathtools}
\usepackage{amsthm}
\usepackage{multirow}
\usepackage{graphicx}
\theoremstyle{plain}
\newtheorem{theorem}{Theorem}[section]

\theoremstyle{definition}
\newtheorem{definition}[theorem]{Definition}

\theoremstyle{remark}

\usepackage{amsthm}

\usepackage[pagebackref,breaklinks,colorlinks]{hyperref}
\usepackage[capitalize,noabbrev]{cleveref}

\usepackage[capitalize]{cleveref}
\crefname{section}{Sec.}{Secs.}
\Crefname{section}{Section}{Sections}
\Crefname{table}{Table}{Tables}
\crefname{table}{Tab.}{Tabs.}

\def\cvprPaperID{40} % *** Enter the CVPR Paper ID here
\def\confName{CVPR}
\def\confYear{2023}

\begin{document}

\title{MASIL: Towards Maximum Separable Class Representation for Few Shot Class Incremental Learning}

\author{Anant Khandelwal\\
Applied Scientist, Amazon
}
\maketitle

%%%%%%%%% ABSTRACT

\begin{abstract}
Few Shot Class Incremental Learning (FSCIL) with few examples per class for each incremental session is the realistic setting of continual learning since obtaining large number of annotated samples is not feasible and cost effective. We present the framework MASIL as a step towards learning the maximal separable classifier. It addresses the common problem i.e forgetting of old classes and over-fitting to novel classes by learning the classifier weights to be maximally separable between classes forming a simplex Equiangular Tight Frame. We propose the idea of concept factorization explaining the collapsed features for base session classes in terms of concept basis and use these to induce classifier simplex for few shot classes. We further adds fine tuning to reduce any error occurred during factorization and train the classifier jointly on base and novel classes without retaining any base class samples in memory. Experimental results on miniImageNet, CIFAR-100 and CUB-200 demonstrate that MASIL outperforms all the benchmarks.
\end{abstract}

\section{Introduction}
\label{submission}
The success of Convolutional Neural Networks (CNN) in wide range of computer vision tasks \cite{NIPS2012_c399862d, he2016deep, ren2015faster, liu2017sphereface, ma2019bayesian, li2020infrared} relies on the fact that the training requires large scale image datasets \cite{deng2009imagenet} and the train and test distributions are almost identical \cite{krizhevsky2017imagenet}. However, when deploying them in real world environments it requires that these models to quickly adapt to changing streams of data and hence can recognize the novel classes emerged over a period of time. But the underlying bottleneck for this adaptation is that CNN requires large amount of data to be collected for each of the novel classes, this takes lot of human effort to annotate them which is infeasible. However, annotating only a few samples seems reasonable, we term this ability to adapt to novel classes (with only few examples) without forgetting the old classes as the \textit{few shot class incremental learning} (FSCIL). Fine-tuning the pre-trained network with limited number of training examples of only novel classes cause the model to forget old classes (\textit{catastrophic forgetting}) and overfitting on recent novel classes \cite{french1999catastrophic, huang2022resolving, tao2020few, dong2021few}. Large amount of studies has been conducted to solve the problem of \textit{catastrophic forgetting} \cite{goodfellow2013empirical}. This includes approaches based on: constraining the weight changes \cite{kirkpatrick2017overcoming, zenke2017continual, dhar2019learning, aljundi2018memory, li2017learning}, retaining the samples from previous data in a memory \cite{parisi2019continual, de2019continual, castro2018end, shin2017continual, aljundi2018memory} data augmentation \cite{yu2020semantic, zhu2021prototype, zhu2022self, xiang2019incremental}, dynamic expansion based architectures (DEA) which expands the network for each new incoming task id while the weights of the base network are frozen for learning keeping both old and new information \cite{fernando2017pathnet, golkar2019continual, hung2019compacting, yan2021dynamically, douillard2020podnet, li2021preserving}. All these approaches are broadly categorised into two main themes i.e. multi-task and multi-class. Multi-task approaches like DEA requires resolving the task id during inference, which is typically unavailable. Multi-class scenario refers to learning a single classifier with a aim to recognize the base and novel classes in a single task. In this paper we study the FSCIL problem under multi-class scenario, since it is more realistic and practical. Recent approaches \cite{zhang2021few, hersche2022constrained, akyurek2021subspace} have proposed to learn the backbone network as feature extractor using data of base classes, and then use this frozen feature extractor to learn the classifier prototypes for novel classes incrementally. But this does not guarantee the maximum separability between the classifier prototypes for base and novel classes and hence can lead to the confusion between the old and new classes resulting in limited performance. Other approaches \cite{chen2020incremental, akyurek2021subspace, hersche2022constrained}, which uses custom loss functions and regularizers to learn the classifier prototypes for novel classes along with preventing \textit{forgetting} on base classes, are also limited by performance because of misalignment between fixed features of base classes and classifier. Recent work \textit{NC-FSCIL}\cite{yang2023neural} proposed the use of neural collapse to learn the maximally separable classifier. They proposed to learn the two layer classifier with pre-fixed simplex weights for base and each incremental sessions. The two layer classifier training with pre-fixed weights for few shot novel classes will not be able to generalize well and hence results in overfitted class representation. Opposed to them we proposed to learn itself the maximum separable class representation using Neural Collapse properties, but we used concept factorization on backbone network to be able to represent any class in general and hence obtain the generalized classifier for novel few shot classes.

In this work, we address this problem of misalignment between fixed features of backbone network (feature extractor) and classifier prototypes to prevent forgetting of base class. Towards that we attempt to learn the maximal separable classifier to avoid confusion between base and novel classes in each incremental session. Our work is inspired from two main studies: 1) Neural Collapse (NC) for imbalanced data \cite{dang2023neural, papyan2020prevalence} and, 2) Concept Factorization \cite{fel2022craft, kalayeh2014nmf}. Neural collapse is the phenomenon where the network when trained beyond zero error towards zero loss, results in collapsing the last layer features of backbone network to form an Equiangular Tight Frame (ETF). The vertices of this frame denotes the feature vector representing the class and aligned with classifier prototype of the corresponding class \cite{papyan2020prevalence}. This guarantees a maximal separable classifier since ETF is a geometric structure forming a simplex where the within class variance is minimized (because of collapse to a single vector) and between class variance is maximized lying at equal angles from each other. However, with pre-fixed simplex for base classes, the feature extractor is easy to train since the sufficient data is available for each of the class in base session ($t=0$) resulting in collapse but for any incremental session $t \ge 1$ with few labelled samples learning the collapsed features for novel classes is challenging since with few samples (as much as 5 samples for a class) the fixed feature extractor is not able to align well with the novel class prototype. To resolve that we introduce the mechanism of concept factorization, where we dissect the collapsed feature extractor on base session to identify the concept basis in the input images. Once the concept basis ("\textit{concept bank}") is identified from base session, we recognize them as the building block from which the incremental session classifier simplex is induced and hence the new set of coefficients can be learnt for inducing the simplex with novel classes. This is additionally fine-tuned along with base class simplex to further align this with few shot instances to reduce any irreducible error occurred during calculating optimal coefficients for the "\textit{concept bank}". This has been illustrated in Fig.\ref{fig:masil}.
To summarize, our main contributions are as follows:
\begin{itemize}
    \item We introduce a novel framework \textbf{MASIL} as an attempt to learn the maximal separable classifier for FSCIL. 
    \item We identified the mechanism where the base session collapsed features (obtained as per Neural Collapse properties) can further be dissected in terms of "\textit{concept bank}", which forms the basis for building classifier prototype of novel classes encountered during incremental session.
    \item Evaluation on three popular FSCIL benchmarks datasets demonstrating state-of-the-art performance. Extensive ablation study has been done to analyze the importance of loss function introduced using Neural collapse properties and the advantage of simplex fine-tuning to reduce the irreducible error. 
\end{itemize}

\section{Related Work} 
\subsection{Few Shot Learning} The idea of few shot learning (FSL) is to adapt the model on novel classes (with only few labelled instances) without caring for the performance on base classes. Most of the works uses meta-learning \cite{ sung2018learning, finn2017model, sun2019meta, snell2017prototypical} or metric learning \cite{snell2017prototypical, vinyals2016matching, sung2018learning}. Recently, the approaches \cite{gidaris2018dynamic, ren2019incremental} have demonstrated the use of meta learning to recognize the base and novel classes both, by sampling "fake" few shot classification task from base classes to learn a classifier for novel classes. 
Finally, the learned classifier weights are combined to jointly recognize the base and novel classes. Some of the works \cite{ren2019incremental} regard this as sort of incremental learning. Contrastively, FSCIL setting is much more realistic where the base dataset is not accessible during the incremental stage and we have to adapt the model for novel classes without catastrophic forgetting \cite{tao2020few, dong2021few}. Metric learning approaches focus on learning a strong backbone network for learning transferable features across the tasks, on top which the similarity function (like k-nearest neighbours in \cite{vinyals2016matching}, non linear distance metric in \cite{sung2018learning} ) is learnt to demonstrate the ability to classify the novel classes with transferable features. However, this requires to train the as much similarity function as the number of incremental sessions in FSCIL but the aim of FSCIL is to train one unified classifier for the base and novel classes. We will discuss in the next section how existing works have dealt the problem of FSCIL different from FSL. 

\begin{figure*}
    \centering
    \includegraphics[width=0.8\textwidth]{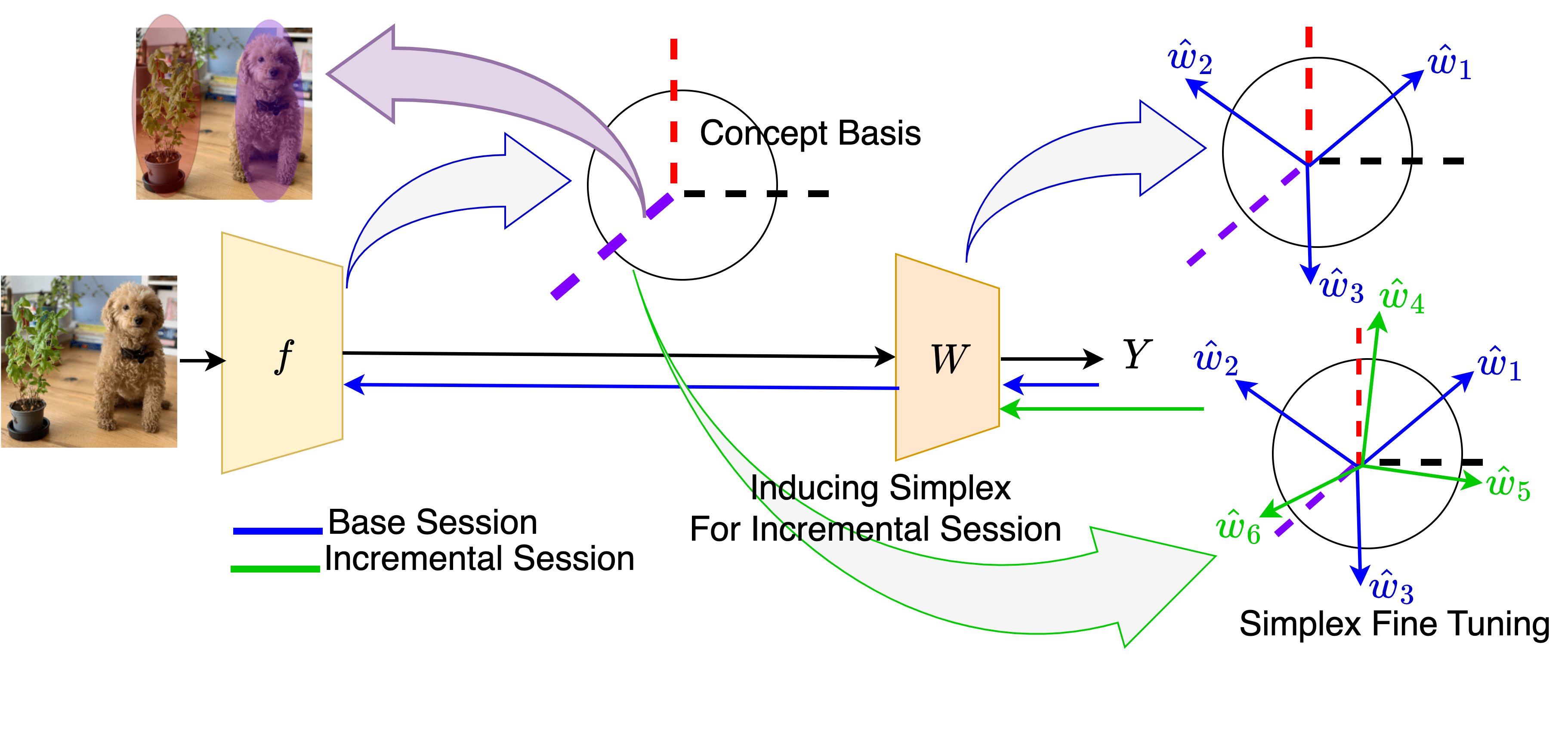}
    \caption{\textbf{\textit{MASIL}}: Illustrating the concept basis obtained from concept factorization of features after feature extractor and their relation with input images. After base session training, classifier simplex for novel classes are induced from these concept basis. Implicit memory (not shown) storing the feature mean of classes seen till current session. This is used to jointly updating the weights of classifier for both base and novel classes during Simplex Finetuning.}
    \label{fig:masil}
\end{figure*}

\subsection{Few Shot Class Incremental Learning}
\textbf{Class Incremental Learning (CIL)}: We start by first discussing the idea of Class Incremental Learning (CIL), it aims to learn a classifier that manages to continuously update itself to recognize all the novel classes without forgetting the base classes \cite{rebuffi2017icarl, cauwenberghs2000incremental, li2017learning}. To overcome this forgetting mechanism CIL studies has been categorized into three broad categories: regularization based \cite{kirkpatrick2017overcoming, li2017learning, dhar2019learning}, rehearsal based \cite{parisi2019continual, de2019continual, aljundi2018memory, castro2018end}, and knowledge distillation \cite{rebuffi2017icarl, hou2019learning, wu2019large}. Regularization based methods constraint the weight changes on the novel classes thereby keeping the information previously learnt for base classes, this causes these methods to suffer for generalization on novel classes because of small allowed change in weights. Rehearsal based methods in which the model is continually be refreshed using old data reserve so that it maintains to learn the novel classes along with old classes. This is limited by the amount of old data it can retain in memory, and the how the instances from old data to be selected for maximal information with minimum memory requirements. These methods are not scalable to large number of classes because of the limited memory. For example, iCaRL\cite{rebuffi2017icarl} learns the nearest neighbour classifier for novel classes while maintaining the memory of exemplars from base session. Knowledge Distillation based methods requires the use of large teacher model to guide the learning of small student model \cite{yang2022rd, hinton2015distilling}. It works by distilling the previously learned information to new model with novel classes, so as to avoid forgetting of base classes. Recent studies \cite{dhar2019learning, douillard2020podnet, hou2019learning} conducted the distillation on feature level rather than on output logit level at the classifier. However, these solutions suffer from a problem of distinguishing between base and novel classes leading to limitation in the performance. 

\textbf{Few Shot Class Incremental Learning (FSCIL)}: Compared to CIL setting, FSCIL aims to learns the novel classes (along with base classes) with few labelled instances \cite{tao2020few, dong2021few}, which is much more realistic and hard, since learning from few instances of novel classes causes over-fitting on novel classes \cite{snell2017prototypical, sung2018learning}. In order to do this, some studies have focused to align base and incremental session using augmentation \cite{peng2022few}, searching for flat minima \cite{shi2021overcoming}. However, for avoiding over-fitting to novel classes it is required that the classifier prototypes for novel classes should be maximally separable from base classes. Adjusting prototypes for base classes is not feasible since that requires the use of base session data. However, these studies \cite{zhang2021few, zhu2021self} have focused on evolving prototypes for novel classes. Large number of existing works have focused on building the custom loss and regularizer \cite{ren2019incremental, hou2019learning, tao2020topology, joseph2022energy, lu2022geometer, hersche2022constrained, akyurek2021subspace, yang2022rd}. However, the same disadvantages we discussed in CIL for regularization and custom loss functions applies in FSCIL as well. In this work we focused on the optimal evolution of prototypes for novel classes which is derived from the same function using which base class prototypes have been developed and ensuring the maximal separability between old and novel classes.

\section{Problem Statement and Context}
In this section we will introduce the problem definition of Few Shot Class Incremental Learning in Section \ref{ps:fscil} and context in subsequent sections.
\subsection{Few Shot Class Incremental Learning}
\label{ps:fscil}
Formally, we define Few Shot Class Incremental Learning (FSCIL) as the stream of labelled data in time sequence as $\mathcal{D}_0$, $\mathcal{D}_1$, ....., where $\mathcal{D}_t = \{(\mathbf{x}_j^t, y_j^t)\}_{j=1}^{j=|\mathcal{D}_t|}$. $\mathcal{C}_t$ be the number of classes in training set $\mathcal{D}_t$, where $ \forall (i, j) \textrm{  } \mathcal{C}_i \cap \mathcal{C}_j = \emptyset$. Specifically, we consider $\mathcal{D}_0$ as the base session with large label space $\mathcal{C}_0$ with each class $c \in \mathcal{C}_0$ have sufficient training images. For $t > 0$ each of the incremental session $\mathcal{D}_t$ have only few labelled images for each novel classes. FSCIL is defined as the time step incremental training of model $\Theta$ on $\mathcal{D}_t \textrm{ } \forall \textrm{ } t > 0$ with no access to any of the previous labelled set from $\mathcal{D}_0$ to $\mathcal{D}_{t-1}$. For $t > 0 \textrm{  } \mathcal{D}_t$ we denote the setting as  C classes with K training examples per class as C-way K-Shot FSCIL where $\mathcal{C}_t \cap \mathcal{C}_t' = \emptyset \textrm{ } \forall t \neq t'$. After each incremental session training with $\mathcal{D}_t$, model $\Theta$ is evaluated to recognize all the training classes encountered so far i.e. $\cup_{i=0}^{i=t} \textrm{ } \mathcal{C}^i$. Hence, FSCIL not only aims to recognize novel classes but to avoid forgetting of the old classes and the setting of learning of novel classes is highly imbalanced and suffers from data scarcity problem as well. This makes FSCIL setting more suited for real world applications.

\textbf{Initialization}: Assuming $\mathcal{C}_0$ as the number of base classes and we have total $T$ incremental session and each session has $k$ classes, so there will be total $K = \mathcal{C}_0 + Tk$ classes. To be able to perform FSCIL, we denote the model trained on base session consists of backbone feature extractor $f(;\theta_f)$ and classifier parameters $\mathbf{W} \in \mathbb{R}^{K \times d}$, where $\mathbf{W}$ is a MLP classifier consisting of L layers denoted as $\mathbf{W} = \mathbf{W}_1 \mathbf{W}_2 ..... \mathbf{W}_L$. For input $X \in \mathbb{R}^{n}$ we denote the features obtained from feature extractor as $\mathbf{H} =  f(X, \theta_f) \in \mathbb{R}^{d \times N}$, where $N$ is the total number of training instances. Similar to \cite{dang2023neural} we also consider last layer features $\mathbf{H}$ as freely optimization variables. The optimization objective is then defined as follows:
\begin{multline}
        \underset{\mathbf{W}, \mathbf{H}}{min} \textrm{ } \mathcal{L}(\mathbf{W}, \mathbf{H}) = \frac{1}{2N}||\mathbf{W}\mathbf{H} - \mathbf{Y}||_F^2 + \frac{\lambda_{\mathbf{W}}}{2}||\mathbf{W}||_F^2 + \\ \frac{\lambda_{\mathbf{H}}}{2}||\mathbf{H}||_F^2
\label{obj}
\end{multline}
where $\mathbf{Y} \in \mathbb{R}^{K \times N}$, is the class label for each of the training instance in $N$ instances and $\lambda_W$, $\lambda_H$ are regularization hyperparameters.

\subsection{Neural Collapse}
\label{ps:nc}
In recent works \cite{papyan2020prevalence, dang2023neural} which have studied the practice of training the DNN beyond zero error towards zero loss. This reveals the geometric structure in the form of simplex equiangular tight frame, formed by the last layer features along with classifier weights. This has been demonstrated on the balanced data and models with various popular architectures. Neural Collapse as defined in \cite{papyan2020prevalence} consists of the following four properties:
\begin{itemize}
    \item ($\mathcal{NC}1$) \textbf{Variability Collapse}: Last layer features of the backbone network for a particular class collapse  to within-class mean. 
    \item ($\mathcal{NC}2$) \textbf{Convergence}: results in optimal class-means which are equally and maximally pairwise separated forming a simplex Equiangular Tight Frame (ETF).
    \item ($\mathcal{NC}3$) \textbf{Classifier Convergence}: Optimal class means forming ETF are aligned to the corresponding classifier weights uptio rescaling. 
    \item ($\mathcal{NC}4$) \textbf{Simplification to nearest class center} When ($\mathcal{NC}1$)-($\mathcal{NC}3$) holds, the model prediction using logits respects nearest class centers.
\end{itemize}
In addition to the balanced data, this \cite{dang2023neural} have derived the geometrical analysis for the imbalanced data given as:
\begin{definition}
\label{def:inj}
\textit{Let ($\mathbf{W}$*, $\mathbf{H}$*) be the global optimizer of equation \ref{obj}, $r = min(K, d)$ and $\mathbf{W} = U_WS_WV_W^T$ be the SVD factorization of $\mathbf{W}$. Then the following holds for the imbalanced data:
\begin{itemize}
    \item ($\mathcal{NC}$1) leads to collapse of features within the same class $\mathbf{H}^{*} = \overline{\mathbf{H}}^{*}\mathbf{Y}$, where $\overline{\mathbf{H}}^{*} =  [\mathbf{h}^{*}_1, \mathbf{h}^{*}_2, .... \mathbf{h}^{*}_K] \in \mathbb{R}^{d \times K}$
    \item ($\mathcal{NC}$3) leads to alignment between classifier weights and corresponding class mean as $\mathbf{w}_k^{*} = \sqrt{\frac{n_k\lambda_H}{\lambda_W}} \mathbf{h}_k^{*} \textrm{ } \forall k \in [K]$, where $n_k$ is the number of instances of class $k$.
    \item ($\mathcal{NC}$2) leads to optimal class means equally and maximally separated forming simplex Equiangular Tight Frame (ETF) $\mathbf{W}^{*}\mathbf{W}^{*\top} = \textrm{ diag }\{s_k^2\}_{k=1}^{k=K}$, where $s_k$ are the singular values of $\mathbf{W}^{*}$
\end{itemize}}
\end{definition}
Another approach i.e. Deep Simplex Classifier\cite{cevikalp2022deep}, proposed the optimization problem as minimization of features obtained from feature extractor to the vertices of simplex as:
\begin{equation}
    \underset{\mathbf{h}_i \in \mathbf{H}^T}{\textrm{min }} \frac{1}{n}\sum_{i=1}^{n} ||\mathbf{h}_i - {s_{y}}_{i}||^2
\label{eqn:ds}
\end{equation}
where ${s_{y}}_{i}$ is vertex of simplex and is treated as the class center for class $y_i$.

\section{MASIL}
Overall framework of our proposed method is illustrated in Fig. \ref{fig:masil}. FSCIL aims to learn the classifier weights $W^{K \times d}$ which works for all classes irrespective of whether they belongs to the base classes during $t=0$ or few shot classes during $t > 0$. Traditionally, this has been achieved by first learning the classifier weights for $\mathcal{C}_0$ base classes and then learn the weights for novel classes $W^{(t)} \in \mathbb{R}^{k \times d}$ with the regularized constraint in the loss function that the old weights $W \in \mathbb{R}^{C_0 + (t-1)k} $ be preserved with little or no updates. However, this leads to misalignment between the classifier prototypes of old and novel classes causing old new confusion(ONC) \cite{huang2022resolving} and catastrophic forgetting \cite{goodfellow2013empirical}. This causes drop in performance of FSCIL classifier as the number of incremental session grows resulting in poor generalizability even in recognizing the base classes. To mitigate this, in this work we adopted the properties obtained from Neural Collapse to learn maximally separable classifier along with concept factorization to learn classifier weights (organized as simplex) for novel classes with few samples. We restricted the feature extractor from updates during incremental session training and rely on concept factorization of the activations obtained for base classes to obtain the basis of concepts called "\textit{concept bank}", using which we can represent maximally separable classifier weights i.e. simplex for few shot classes. To represent the classifier simplex using "\textit{concept bank}" it requires to solve only for the coefficient matrix which can be done by just solving the Non Negative Least Squares (NNLS).

\subsection{Concept Factorization}
\label{conc}
The idea of concept factorization relates to the phenomena of neural collapse, where it learns to maximally separate the classes by forming the simplex at class level on both levels of class features and classifier weights. In order to achieve this it merge the activations (during forward pass) of the same class until they all converge to the one hot class vector at the logits layer as depicted in equation \ref{eqn:ds}. This allows the class wise feature vectors which are concentrated at higher layers to be recursively broken into multiple concepts moving from highest layer to lower layers tracing back to the input images where it can be explained with regions as concepts, combination of which makes it possible to be able to classify it to particular class. We adopted NMF (Non Negative Matrix Factorization) as in \cite{fel2022craft} of activations obtained at the output of feature extractor given as:
\begin{equation}
    \underset{\mathbf{P} \ge 0, \mathbf{Q} \ge 0}{\textrm{min }} \frac{1}{2} ||\mathbf{A} - \mathbf{P}\mathbf{Q}^T||_F^2
\label{nmf}
\end{equation}
where, $||.||_F$ is the Frobenius norm, the activations $\mathbf{A} \in \mathbb{R}^{n \times d}$ obtained from crop of images $X_i = {\tau}(x_i) \textrm{  } X_i \in X^{n \times p}$ with $\tau$ is a crop function. We take random crops (governed by $\tau$) of images, this results in unique concepts across the categories to be able to build the bank of unique concept vectors called "\textit{concept bank}". Activations at the last layer of feature extractor after global pooling for these random crops is given as $\textbf{A} = f(X, \theta_f) \in \mathbb{R}^{n \times d}$. NMF is simply the factorization of concept activations $\mathbf{A}$ into the "\textit{concept bank}" $\mathbf{Q} \in \mathbb{R}^{v \times d}$ (where it follows low rank factorization $v \ll \mathbf{min}(n, p)$) and coefficients $\mathbf{P} \in \mathbb{R}^{n \times v}$ denote the importance of each of the concepts in explaining the activations $\mathbf{A}$. Once the "\textit{concept bank}" is precomputed, we can obtain the coefficients $P(x)$ for any input $x$ using NNLS (Non-Negative Least Squares) i.e. $\underset{P \ge 0}{min} \textrm{  } \frac{1}{2} ||f(x;\theta_f) - P(x)\mathbf{Q}^T||_F^2$. Relating activation factorization in equation \ref{nmf} and neural collapse in equation \ref{eqn:ds}, implies that the activations when collapses to the mean features vector for each class forming the class simplex vector which is composed of concept basis vectors and the corresponding coefficients, combining for all classes which gives the overall basis called "\textit{concept bank}". 

\subsection{NMF Layer}
During NMF factorization of equation \ref{nmf}, we keep the feature extractor $f(.;\theta_f)$ frozen. We approached the NMF problem solution using \textbf{ADMM} (Alternating Direction Method of Multipliers) \cite{boyd2011distributed} since NMF is non-convex, but however it can be made convex by fixing the value of either of the two factors $(\mathbf{P, Q})$ which requires alternating update of either of two factors fixing one at a time, which is equivalent to solving a Non-Negative Least Squares (NNLS) problem making it convex. This alternating update mechanism called as \textbf{ADMM}, formulated as:
\begin{equation}
    \mathbf{P}_{t + 1} = \textrm{arg }\underset{\mathbf{P} \ge 0}{\textrm{min }} \frac{1}{2} ||\mathbf{A - PQ_t}^T||_F^2
\end{equation}
\begin{equation}
    \mathbf{Q}_{t + 1} = \textrm{arg }\underset{\mathbf{Q} \ge 0}{\textrm{min }} \frac{1}{2} ||\mathbf{A - P_t Q}^T||_F^2
    \label{cb}
\end{equation}
It ensures global or local minimum since each of the NNLS problem obeys \textit{Karush–Kuhn–Tucker} (KKT) optimality conditions \cite{karush2014minima, kuhn2014nonlinear}. Using these conditions forming the implicit function \cite{griewank2008evaluating} makes the implicit differentiation \cite{griewank2008evaluating, krantz2002implicit, bell2008algorithmic} allows to compute the gradients $(\frac{\partial P}{\partial A}, \frac{\partial Q}{\partial A})$, but however we have to relate the concepts with the input image regions we require to compute $(\frac{\partial P}{\partial X}, \frac{\partial Q}{\partial X})$. This can be calculated as:
\begin{equation}
    \frac{\partial P}{\partial X} = \frac{\partial A}{\partial X} \frac{\partial P}{\partial A}, \textrm{  
  } \frac{\partial Q}{\partial X} = \frac{\partial A}{\partial X} \frac{\partial Q}{\partial A}
  \label{del}
\end{equation}
Computation of $\frac{\partial A}{\partial X}$ is fairly straight word using \textit{Pytorch}. More details on implementation of combining gradients from implicit differentiation in \textit{Jax} \cite{huang2016flexible, blondel2021efficient} and gradient from \textit{Pytorch} computation is detailed in Section \ref{app:b}. Once we precompute the "concept bank" for base classes using equation \ref{cb} and \ref{del}, we fixed the $\mathbf{Q}$ and allows only to compute optimal coefficients $P(x)$ for any input $x$ using NNLS 
\begin{equation}
    \underset{\mathbf{P} \ge 0}{\textrm{min }} \textrm{  } \frac{1}{2} ||f(x;\theta_f) - P(x)\mathbf{Q}^T||_F^2
\label{inf}
\end{equation}
which give the optimal representation of activation for any input $x$ in terms concept basis vectors. 

\subsection{Classifier Simplex Representation}
\label{nsc}
Equation \ref{eqn:ds} is optimizing the feature representation for each class resulting in collapsed representation for class $y_i$ as $s_{y_i}$. Similarly, equation \ref{obj} results in simplex representation for each class i.e. $\mathbf{w}_{y_i} \in \mathbf{W}$. So if we consider the normalized simplex representation on a unit hypersphere \cite{cevikalp2022deep} of each class then:
\begin{equation}
     \mathbf{w}_{y_i}^T s_{y_i} = 1 \textrm{  } \forall \textrm{  } y_i \in \cup_{j=0}^{j=t}\mathcal{C}_j
\label{ws}
\end{equation}
which results in the modified loss function of equation \ref{eqn:ds} to:
\begin{equation}
    \underset{\mathbf{h}_i}{\textrm{min }}\frac{1}{|\mathcal{D}_j|} \sum_{(x_i, y_i) \in \mathcal{D}_j}||\mathbf{w}_{y_i}^T\mathbf{h}_i - 1||_F^2 
\label{loss}
\end{equation}
s.t. $\mathbf{w}_{y_i}^T s_{y_i} = 1$ which is same as in equation \ref{obj} and hence follow the neural collapse properties. Moreover, optimizing equation \ref{eqn:ds}, results in the collapsed feature representation for all instances belonging to that class. Additionally, equation \ref{nmf} computes the best approximation of collapsed feature representation $\mathbf{H} \approx \mathbf{PQ}^T$. For any input $(x_i, y_i)$ belongs to $\mathcal{D}_j, \textrm{ } j > 0$, then optimal $\mathbf{h}_{i}$ obtained from equation \ref{inf} is given as:
\begin{equation}
    \mathbf{h}_i = \mathbf{P}(x_i)\mathbf{Q}^T
\label{hi}
\end{equation}
From ($\mathcal{NC}1$), the collapsed feature representation of each class converge to a unique vector e.g. for class $y_i$ the feature representation of all instances is denoted as $\mathbf{H}_{y_i} \in \mathbf{H}$, ($\mathcal{NC}1$) implies covariance $\sum_{\mathbf{H}_{y_i}} \xrightarrow{} 0$. i.e. the features collapse to their corresponding class means i.e. $\mathbf{h}_{y_i}^{*} = \sum_{i=1}^{n_{y_i}} \mathbf{h}_i$, where $n_{y_i}$ is the number of instances for class $y_i$, and as per loss in equation \ref{eqn:ds}, this is minimum when $s_{y_i} = \mathbf{h}_{y_i}^{*}$, then from equation \ref{ws} and \ref{hi}:
\begin{align}
    \mathbf{\hat{w}}_{y_i} = \frac{1}{|\mathcal{D}_j|} \left ( \sum_{(x_i, y_i) \in \mathcal{D}_j}\mathbf{P}(x_i)  \right )\mathbf{Q}^T \textrm{    } \forall j > 0
\label{weights}
\end{align}
where the coefficients $\mathbf{P}(x_i)$ for each instance of class $y_i$ are calculated using NNLS as per equation \ref{inf}, additionally, $y_i \in \mathcal{C}_j, j > 0$ are the few shot classes and classifier weights are the optimal simplex representation for few shot classes. For base session classes ($j=0$) the classifier simplex representation is simply $\mathbf{\hat{w}}_{y_i} = s_{y_i}^T$. Since we implemented the classifier using MLP with $L=2$ layers, for each layer the simplex representation is $\mathbf{\hat{w}}_{l, y_i} = (\mathbf{\hat{w}}_{y_i})^{1/L}$.

\subsection{Simplex Finetuning}
\label{csf}
In Section \ref{nsc} we described the optimal simplex representation for each class belongs to the few shot class $y_i \in \mathcal{C}_j, j > 0$. But, however due to the the inherent irreducible error to NNLS, we approach the optimal representation of simplex for few shot class by further fine-tuning the classifier weights (keeping the feature extractor frozen) initialized using simplex representation as obtained in equation \ref{weights}. To avoid deviating the weights to much from optimal simplex representation we add a constraint to the loss in equation \ref{loss} as:
\begin{multline}
  \underset{\mathbf{w}_{y_i}}{\textrm{min}}  \mathcal{L}(\mathbf{w}_{y_i}) = \frac{1}{|\mathcal{D}_j|} \sum_{(x_i, y_i) \in \mathcal{D}_j}||\mathbf{w}_{y_i}^T\mathbf{h}_i - 1||_F^2 + \\ \alpha ||\mathbf{w}_{y_i} - \mathbf{\hat{w}}_{y_i}||_F^2, \alpha \in [0,1]
  \label{add_loss}
\end{multline}
where $y_i \in \mathcal{C}_j, j > 0$ and feature extractor is frozen and hence optimizing for the best $\mathbf{w}_{y_i}$. Since base session training (i.e. $j=0 \textrm{ and dataset } \mathcal{D}_0$) is governed using the loss function of equation \ref{eqn:ds} and hence results in collapsed representation of features at the terminal layer for each class $y_i \in \mathcal{C}_0$. For the simplex representation for each class in $\mathcal{C}_0$ to remain maximally separable with the ones obtained for few shot class we utilized the collapsed representation of features for each class to further fine tune the simplex representation, but without keeping the image instances in memory we memorized the collapsed representation (which is the mean representation of instance features for each class) in $\mathcal{M}$ given as:
\begin{equation}
    \mathcal{M}_{y_i} = \frac{1}{n_{y_i}} \sum_{i=1}^{n_{y_i}} \mathbf{h}_i, \forall \mathcal{M}_{y_i} \in \mathcal{M}
\label{mem}
\end{equation}
where, $n_{y_i}$ is the number of instances of class $y_i$. The updated loss function during fine tuning stage include the base session classes and few shot class is given as:
\begin{multline}
      \underset{\mathbf{w}_{y_i}}{\textrm{min}}  \mathcal{L}(\mathbf{w}_{y_i}) = \frac{1}{|\mathcal{D}_j|} \sum_{(x_i, y_i) \in \mathcal{D}_j}||\mathbf{w}_{y_i}^T\mathbf{h}_i - 1||_F^2 + \\ \frac{1}{|\mathcal{M}|} \sum_{(\mathcal{M}_{y_i}, y_i) \in \mathcal{M}}||\mathbf{w}_{y_i}^T\mathcal{M}_{y_i} - 1||_F^2 + \\ \alpha ||\mathbf{w}_{y_i} - \mathbf{\hat{w}}_{y_i}||_F^2, \alpha \in [0,1] 
\label{final_loss}
\end{multline}

\begin{table*}
\centering
\scriptsize
\begin{tabular}{llllllllllll}
\toprule
\multicolumn{1}{c}{\multirow{2}{*}{\textbf{Methods}}} & \multicolumn{9}{c}{\textbf{Session Accuracy (\%) ($\uparrow$)}}                                       & \multicolumn{1}{c}{\multirow{2}{*}{\textbf{\begin{tabular}[c]{@{}c@{}}Average \\ Acc. ($\uparrow$)\end{tabular}}}} & \multicolumn{1}{c}{\multirow{2}{*}{\textbf{\begin{tabular}[c]{@{}c@{}}Relative \\ Improvement\end{tabular}}}} \\ \cmidrule{2-10}
\multicolumn{1}{c}{}                                  & \multicolumn{1}{c}{\textbf{0}} & \multicolumn{1}{c}{\textbf{1}} & \multicolumn{1}{c}{\textbf{2}} & \multicolumn{1}{c}{\textbf{3}} & \multicolumn{1}{c}{\textbf{4}} & \multicolumn{1}{c}{\textbf{5}} & \multicolumn{1}{c}{\textbf{6}} & \multicolumn{1}{c}{\textbf{7}} & \multicolumn{1}{c}{\textbf{8}} & \multicolumn{1}{c}{}                                                                                                            & \multicolumn{1}{c}{}                                                                                          \\ \midrule
iCaRL \cite{rebuffi2017icarl}                                               &      61.31      &    46.32        &     42.94       &     37.63       &     30.49       &   24.00         &    20.89        &    18.80        &   17.21         &       33.29                                                                                                                          &  \textbf{+41.65    }                                                                                                         \\
NCM \cite{hou2019learning}                                                 &    61.31         &   47.80         &        39.30     &    31.90         &    25.70         &    21.40        &     18.70       &     17.20        &    14.17        &      30.83                                                                                                                           &  \textbf{+44.69 }                                                                                                            \\
D-Cosine \cite{vinyals2016matching}                                             &    70.37          &      65.45       &     61.41        &    58.00        &   54.81         &     51.89       &    49.10        &   47.27         &      45.63      &       55.99                                                                                                                          &      \textbf{ +13.23 }                                                                                                       \\ \midrule
TOPIC \cite{tao2020few}                                                &   61.31          &    50.09        &   45.17         &    41.16         &     37.48       &    35.52         &    32.19        &   29.46         &    24.42         &      39.64                                                                                                                           &  \textbf{+34.44 }                                                                                                            \\
IDLVQ \cite{chen2020incremental}                                                &   64.77         &     59.87        &       55.93      &     52.62       &    49.88        &   47.55         &   44.83          &    43.14        &      41.84      &    51.16                                                                                                                               &      \textbf{  +17.02}                                                                                                       \\
Self-promoted \cite{zhu2021self}                                        &    61.45         &     63.80       &          59.53   &   55.53         &    52.50        &   52.50         &     46.69       &  43.79          &    41.92        &   52.76                                                                                                                              &  \textbf{+16.94   }                                                                                                          \\
CEC \cite{zhang2021few}                                                  &  72.00          &   66.83          &   62.97          &   59.43         &    56.70        &    53.73         &    51.19        &   49.24          &    47.63         &     57.75                                                                                                                            &   \textbf{+11.23 }                                                                                                           \\
LIMIT \cite{zhou2022few}                                                &   72.32          &   68.47         &   64.30          &     60.78        &     57.95        &       55.07     &     52.70        &    50.72         &    49.19         &       59.06                                                                                                                          &     \textbf{ +9.67  }                                                                                                       \\
Regularizer \cite{akyurek2021subspace}                                          &   80.37          &    74.68        &      69.39       &     65.51        &    62.38        &      59.03       &      56.36       &    53.95         &       51.73      &    63.71                                                                                                                             &    \textbf{  +7.13  }                                                                                                       \\
MetaFSCIL \cite{chi2022metafscil}                                            &    72.04        &    67.94         &     63.77       &   60.29          &    57.58        &      55.16       &     52.90         &       50.79       &      49.19      &      58.85                                                                                                                         & \textbf{ +9.67   }                                                                                                          \\
C-FSCIL \cite{hersche2022constrained}                                              &   76.40          &     71.14        &     66.46        &     63.29        &   60.42         &    57.46        &    54.78         &    53.11        &    51.41        &     61.61                                                                                                                              &    \textbf{+7.45  }                                                                                                         \\
Data-free Replay \cite{liu2022few}                                     &   71.84         &   67.12          &    63.21         &    59.77         &      57.01       &   53.95         &   51.55          &    49.52         &       48.21     &        58.02                                                                                                                          &   \textbf{+10.65   }                                                                                                         \\
ALICE \cite{peng2022few}                                                &    80.60         &     70.60       &   67.40          &     64.50         &   62.50         &   60.00          &    57.80         &    56.80        &   55.70         &       63.99                                                                                                                          &    \textbf{+3.16}                                                                                                          \\
SSFE-Net \cite{pan2023ssfe}                                             &     72.06        &    66.17        &     62.25        &    59.74        &     56.36        &   53.85         &  51.96          &      49.55      &    47.73        &     57.74                                                                                                                            &   \textbf{+11.13}                                                                                                            \\ 

NC-FSCIL \cite{yang2023neural}                                             &      84.02        &     76.80        &     72.00       &     67.83        &      66.35        &   64.04         &   61.46          &      59.54      &     58.31        &      67.82                                                                                                                            &   \textbf{+0.55}                                                                                                            \\ \midrule
\textit{\textbf{MASIL(Ours)}}                         &    \textbf{85.15 }       &     \textbf{77.00 }      &     \textbf{ 72.20}      &   \textbf{67.92}         &     \textbf{66.60}       &    \textbf{64.2}        &     \textbf{ 61.50}      &    \textbf{59.60 }       &    \textbf{58.86 }       &       \textbf{68.11}                                                                                                                          &                                                                                                               \\ \bottomrule
\end{tabular}
\caption{Performance comparison on miniImageNet with ResNet-18 as backbone architecture under \textit{5-way 5-shot} FSCIL setting. Table denotes the accuracy in each session, average accuracy across sessions and "Relative Improvement" denotes the improvement of our method in the last session. Methods above separating line are CIL methods for FSCIL as in \cite{tao2020few} and \cite{zhang2021few}}
\label{tab:1}
\end{table*}

\begin{table*}[htb]
\centering
\scriptsize
\begin{tabular}{llllllllllll}
\toprule
\multicolumn{1}{c}{\multirow{2}{*}{\textbf{Methods}}} & \multicolumn{9}{c}{\textbf{Session Accuracy (\%) ($\uparrow$)}}                                       & \multicolumn{1}{c}{\multirow{2}{*}{\textbf{\begin{tabular}[c]{@{}c@{}}Average \\ Acc. ($\uparrow$)\end{tabular}}}} & \multicolumn{1}{c}{\multirow{2}{*}{\textbf{\begin{tabular}[c]{@{}c@{}}Relative \\ Improvement\end{tabular}}}} \\ \cmidrule{2-10}
\multicolumn{1}{c}{}                                  & \multicolumn{1}{c}{\textbf{0}} & \multicolumn{1}{c}{\textbf{1}} & \multicolumn{1}{c}{\textbf{2}} & \multicolumn{1}{c}{\textbf{3}} & \multicolumn{1}{c}{\textbf{4}} & \multicolumn{1}{c}{\textbf{5}} & \multicolumn{1}{c}{\textbf{6}} & \multicolumn{1}{c}{\textbf{7}} & \multicolumn{1}{c}{\textbf{8}} & \multicolumn{1}{c}{}                                                                                                            & \multicolumn{1}{c}{}                                                                                          \\ \midrule
iCaRL \cite{rebuffi2017icarl}                                               &      64.10      &     53.28      &      41.69       &    34.13      &     27.93       &   25.06         &    20.41        &    15.48       &   13.73         &       32.87                                                                                                                          &  \textbf{+42.42}                                                                                                             \\
NCM \cite{hou2019learning}                                                 &    64.10        &    53.05         &       43.96     &   36.97         &   31.61          &     26.73        &     21.23      &      16.78         &   13.54        &       34.22                                                                                                                           &   \textbf{+42.61}                                                                                                            \\
D-Cosine \cite{vinyals2016matching}                                             &    74.55         &      67.43       &      63.63        &     59.55        &    56.11         &    53.80       &    51.68        &   49.67         &      47.68      &       58.23                                                                                                                         &     \textbf{+8.47}                                                                                                          \\ \midrule
TOPIC \cite{tao2020few}                                                &   64.10         &     55.88       &   47.07         &    45.16         &      40.11      &     36.38         &    33.96        &   31.55         &     29.37        &     42.62                                                                                                                           &  \textbf{26.78}                                                                                                             \\
Self-promoted \cite{chen2020incremental}                                                &   64.10         &      65.86        &       61.36     &    57.45      &    53.69        &    50.75         &    48.58          &     45.66        &      43.25      &    54.52                                                                                                                              &      \textbf{+12.9}                                                                                                         \\
CEC \cite{zhang2021few}                                                  &   73.07          &   68.88          &    65.26          &    61.19         &     58.09      &    55.57        &    53.22         &    51.34          &     49.14         &     59.53                                                                                                                           &   \textbf{+7.01 }                                                                                                           \\
DSN \cite{chi2022metafscil}                                            &     73.00        &    68.83        &    64.82       &  62.64          &     59.36       &      56.96       &      54.04         &       51.57       &      50.00      &       60.14                                                                                                                        & \textbf{+6.15}                                                                                                              \\

LIMIT \cite{zhou2022few}                                                &   73.81          &    72.09         &    67.87          &     63.89        &     60.70        &       57.77     &      55.67       &    53.52         &     51.23        &        61.84                                                                                                                          &     \textbf{+4.92}                                                                                                          \\
MetaFSCIL \cite{akyurek2021subspace}                                          &    74.50           &     70.10        &      66.84    &     62.77        &     59.48       &      56.52       &       54.36       &    52.56         &      49.97       &     60.79                                                                                                                             &    \textbf{+6.18 }                                                                                                          \\

C-FSCIL \cite{hersche2022constrained}                                              &  77.47         &     72.40       &      67.47       &      63.25       &    59.84        &    56.95       &     54.42         &    52.47       &    50.47        &      61.64                                                                                                                             &     \textbf{+5.68 }                                                                                                         \\
Data-free Replay \cite{liu2022few}                                     &    74.40        &    70.20          &     66.54        &    62.51        &      59.71     &   56.58         &    54.52         &    52.39        &       50.14   &        60.78                                                                                                                          &   \textbf{+6.01}                                                                                                            \\
ALICE \cite{peng2022few}                                                &     79.00         &    70.50      &   67.10         &     63.40         &    61.20       &   59.20          &    58.10        &     56.30       &   54.10         &      63.21                                                                                                                          &   \textbf{+2.05}                                                                                                            \\ 

NC-FSCIL \cite{yang2023neural}                                             &      82.52        &     76.82        &    73.34        &    69.68        &     66.19        &    62.85        &   60.96         &       59.02      &   56.11      &     67.50                                                                                                                            &   \textbf{+1.12}                                                                                                            \\ \midrule

\textit{\textbf{MASIL(Ours)}}                         &    \textbf{82.55}        &     \textbf{76.98}      &     \textbf{73.44}       &    \textbf{69.75}        &    \textbf{66.48}        &    \textbf{62.98}        &  \textbf{61.4}          &     \textbf{59.81}       &    \textbf{57.23}        &     \textbf{67.84}                                                                                                                            &                                                                                                               \\ \bottomrule
\end{tabular}
\caption{Performance comparison on CIFAR-100 with ResNet-18 as backbone architecture under \textit{5-way 5-shot} FSCIL setting. Table denotes the accuracy in each session, average accuracy across sessions and "Relative Improvement" denotes the improvement of our method in the last session. Methods above separating line are CIL methods for FSCIL as in \cite{tao2020few} and \cite{zhang2021few}}
\label{tab:2}
\end{table*}

where the constraint is now valid for base session classes as well along with few shot classes with the fact that simplex representation for each class should not deviate much (depends on the contributing factor $\alpha$) from the optimal simplex representation. In each incremental session we train our classifier network using this loss function after deriving the simplex representation for each few shot classes from equation \ref{weights}.

\section{Experiments}
We prove the effectiveness of MASIL on three well known FSCIL benchmark datasets (as in ALICE \cite{peng2022few}) described in Section \ref{app:a} along with FSCIL setting and compared its performance with the state-of-the-art methods (Section \ref{beeval}). Training details and hyper parameters are discussed in Appendix \ref{app:b}. 

\subsection{Dataset Details}
\label{app:a}
\begin{itemize}
    \item \textbf{CIFAR-100} \cite{krizhevsky2009learning} consists of 100 classes in total with color images 
     of size $32 \times 32$. Each class consists of 500 images for training and 100 images for testing. The base session ($t = 0$) consists of 60 classes and the rest 40 classes contributed for 8 incremental session with 5-way 5-shot setting (i.e. 5 images for each of the 5 classes) for $1 \le t \le 8$.
    \item \textbf{miniImageNet} \cite{russakovsky2015imagenet} is a variant of ImageNet\cite{5206848} with color images of size $84 \times 84$. It also consits of same number of classes as CIFAR-100 and same number of images in train and test, resulting in the same configuration for base and incremental sessions. 
    \item \textbf{CUB-200} \cite{wah2011caltech} consists of 11,788 images (size $224 \times 224$) in total spanning across 200 classes. There are 5,994 images in train and 5,794 images in test. Base session ($t=0$) consists of 100 classes and rest 100 classes contributed towards 10 incremental session ($1 \le t \le 10$) with 10-way 5-shot setting (5 images for 10 classes each). 
\end{itemize}

\subsection{Benchmark Evaluation}
\label{beeval}
Performance comparison on miniImageNet, CIFAR-100 and CUB-200 is demonstrated in Table \ref{tab:1}, \ref{tab:2} and \ref{tab:3} (given in Appendix \ref{addres} due to space limitation) respectively. Our method \textbf{\textit{MASIL}} outperforms in all the methods in the last session with relative improvement of +3.16\%, +2.05\% and +0.14\% on miniImageNet, CIFAR-100, CUB-200 respectively as compared to strongest baseline ALICE \cite{peng2022few}. Additionally, our method outperforms all the methods in all the sessions (except on CUB-200 session 2). Moreover, on average accuracy our method outperforms atleast by +1.79\% as compared to strongest baseline, collectively is an indicator that our model helps in mitigating the forgetting issue in a realistic setting of continual learning namely \textit{FSCIL}. \vspace{-2mm} 

\begin{table*}[htb]
\centering
\begin{tabular}{l|ll|ll|ll}
\toprule
\multicolumn{1}{c|}{\multirow{2}{*}{\textbf{Models}}} & \multicolumn{2}{c|}{\textbf{miniImageNet}}               & \multicolumn{2}{c|}{\textbf{CIFAR-100}}                  & \multicolumn{2}{c}{\textbf{CUB-200}}                    \\ \cmidrule{2-7} 
\multicolumn{1}{c|}{}                                 & \multicolumn{1}{c}{Final ($\uparrow$)} & \multicolumn{1}{c|}{Average ($\uparrow$)} & \multicolumn{1}{c}{Final ($\uparrow$)} & \multicolumn{1}{c|}{Average ($\uparrow$)} & \multicolumn{1}{c}{Final ($\uparrow$)} & \multicolumn{1}{c}{Average ($\uparrow$)} \\ \midrule
Learnable + CE                                        &   50.04                        &    61.30                          &    52.13                       &    62.68                          &   50.38                        &  59.58                           \\
$\mathcal{NC}$ + CE                                   &   56.66                        &      68.23                        &   54.42                        &   64.00                            &   56.83                        &     65.51                        \\
$\mathcal{NC}$ + ETF Loss                             &   58.31                        &    67.82                          &   56.11                        &   67.50                           &  59.44                         &    67.28                          \\
$\mathcal{NC}$ + ETF Loss + CF              &    58.72                       &       68.04                       &    56.13                       &         67.51                     &     59.72                      &      67.45                       \\ \midrule
\textit{\textbf{MASIL}(Ours)}        &    58.86                       &      68.11                        &  57.23                         &     67.84                         &     60.24                      &    67.54                         \\ \bottomrule
\end{tabular}
\caption{Ablation Studies on three datasets investigating the effects of Simplex based loss, Concept Factorization and Simplex Fine Tuning}
\label{tab:abl}
\end{table*}
\begin{figure*}[htb]
    \centering
    \begin{minipage}{0.48\linewidth}
        \centering
        \includegraphics[width=0.8\linewidth]{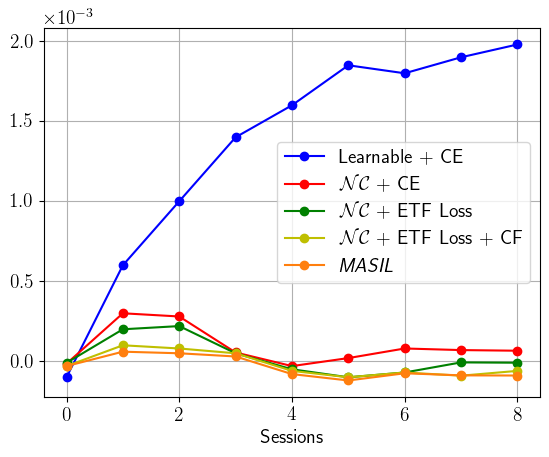}%{c.PNG} % first figure itself
    \end{minipage}\hfill
    \begin{minipage}{0.48\linewidth}
        \centering
        \includegraphics[width=0.8\linewidth]{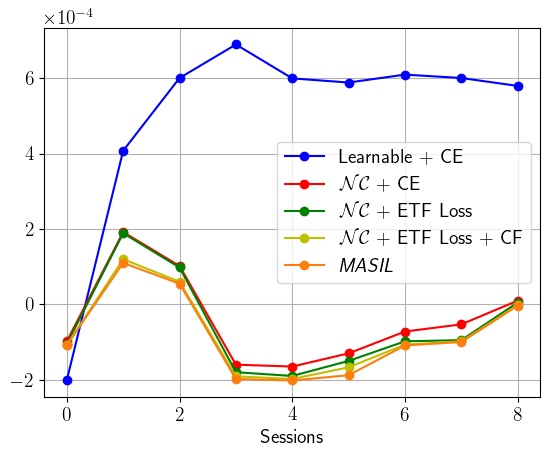}%{h.PNG} % second figure itself
    \end{minipage}
    \caption{Average cosine similarities between different classes at each session for Train (Left) and Test (Right) on miniImageNet. Calculation of cosine similarity is done for all the classes encountered so far after the model gets trained in current session.}
    \label{plot:ana}
    \vspace{-4mm}
\end{figure*}
\subsection{Ablation Studies}
We consider variations to base model (backbone network i.e. ResNet-18 with classifier and memory as introduced in equation \ref{mem}) to validate the 1) effects of loss introduced in equation \ref{loss} (ETF) as compared to cross-entropy (CE) loss with and without neural collapse induced simplex classifier prototypes, 2) effects of few shot simplex induced with concept factorization and 3) effects of simplex fine tuning. To validate the first effect there are two models. The first model (Learnable + CE) uses a classifier with learning weights from CE loss, which is the most common practice. Second model ($\mathcal{NC}$ + CE) uses the CE loss with classifier weights as per the neural collapse properties but uses the CE loss instead of the loss in equation \ref{loss}. To validate the second effect (third model i.e. $\mathcal{NC}$ + ETF Loss) we did not initialize the classifier weights for few shot classes and train them as per the loss in equation \ref{loss} with memory of base classes as in equation \ref{mem}. For third effect i.e. fourth model ($\mathcal{NC}$ + ETF Loss + CF) we reported the performance without fine tuning and just use the classifier weights calculated from concept factorization (CF) as in equation \ref{weights}. Finally, we reported performance of \textbf{\textit{MASIL}} to compare among all of them. As shown in Table \ref{tab:abl}, adopting the loss function in equation \ref{loss} is definitely helps in mitigating performance drop as compared to CE loss even with classifier weights is assumed to be forming simplex, and it further mitigates using the weight initialized with CF and further with fine tuning. It indicates the success of CF along with neural collapse towards optimal solution for FSCIL.

\subsection{Analysing Classifier Weights}
We further analysed the classifier weights alignment with respect to the mean feature (collapsed feature) of each class. We used the classifier weights and mean feature from each of the models described in ablation studies to validate the effect of \textit{MASIL} in learning the maximal separable classifier, where the separable property between classes is measured by cosine similarity. Specifically, we plotted the average cosine similarities between mean feature and the classifier weights of different classes i.e. $\textrm{Avg}_{k \neq k'}\{h_k \cdot w_k'\}$ for both train and test datasets. We have illustrated this for miniImageNet in Fig. \ref{plot:ana}. Clearly, on both the train and test the similarity between different classes goes on increasing for the "Learnable + CE" model. While using the loss in equation \ref{loss} (as per Neural Collapse) have no increasing trend. Incorporating concept factorization and simplex fine tuning (in \textit{MASIL}) further reduces the similarities as the session grows on and hence mitigate the effect of forgetting and confirming the maximum separability with \textit{MASIL}. 

\section{Conclusion}
In this paper we propose the novel framework \textit{MASIL} as an step towards learning the maximum separable classifier in a competitive setting of continual learning i.e. FSCIL. We propose to induce the simplex from concept factorization helps in few shot cases. We introduced novel loss function where the base and novel classes can be learnt together during fine tuning to further mitigate forgetting and overfitting. In experiments \textit{MASIL} outperforms all the benchmarks with sufficient margin on three datasets proving its efficiency.

%%%%%%%%% REFERENCES
{\small
\bibliographystyle{ieee_fullname}
\bibliography{egbib}
}
\newpage

\appendix

\section{Implementation Details}
\label{app:b}
\textbf{Backbone Architecture}: Existing works in FSCIL have leveraged ResNet-18, ResNet-12, ResNet-20 \cite{he2016deep} as the backbone network for feature extractor. Following ALICE \cite{peng2022few}, we use ResNet-18 as the backbone network for feature extractor on top which the two layer MLP for projecting the features as the classification layer is trained for base and incremental sessions. 

\textbf{Concept Factorization}: We used low rank factorization variable $v=64$ for CIFAR and miniImageNet, and $v=72$ for CUB-200. For $\tau$ function, it corresponds to randomly choosing 10 cropped patches of size $18 \times 18$ on CIFAR-100 and miniImageNet and patch size of $64 \times 64$ for CUB-200. We didn't use the \textit{scikit-learn implementation} \cite{pedregosa2011scikit} of NMF, we leverage the work of \cite{fel2022craft, huang2016flexible}, which uses \textit{Jax} \cite{blondel2021efficient, pedregosa2011scikit} implementation of ADMM\cite{boyd2011distributed} using \textit{Jaxopt} library. We convert the \textit{Jax} array to tensor array to be able to combine with the tensor array obtained from Pytorch on the gradient $\frac{\partial A}{\partial X}$ and compute the gradient with respect to input images i.e. $\frac{\partial P}{\partial X}, \frac{\partial Q}{\partial X}$. 

\begin{table}[htb]
\centering
\resizebox{\columnwidth}{!}{%
\begin{tabular}{ccccc}
\toprule
\multicolumn{1}{c|}{\multirow{2}{*}{\textbf{Dataset}}} & \multicolumn{2}{c|}{\textbf{Base Session}}                                                               & \multicolumn{2}{c}{\textbf{Incremental Session}}                                        \\ \cmidrule{2-5} 
\multicolumn{1}{c|}{}                                  & \textbf{Epcohs} & \multicolumn{1}{c|}{\textbf{\begin{tabular}[c]{@{}c@{}}Learning \\ Rate\end{tabular}}} & \textbf{Iterations} & \textbf{\begin{tabular}[c]{@{}c@{}}Learning \\ Rate\end{tabular}} \\ \midrule
CIFAR-100                                              & 200             & 0.25                                                                                   & 20-80               & 0.25                                                              \\
miniImageNet                                           & 500             & 0.25                                                                                   & 120-190             & 0.025                                                             \\
CUB-200                                                & 80              & 0.025                                                                                  & 80-150              & 0.05                                                              \\ \bottomrule
\end{tabular}%
}
\caption{Training Details for Base and Incremental Session}
\label{tab:epc}
\end{table}

\textbf{Training Details}: Data augmentation strategies like random crop, horizontal flip, rotation, brightness variation, cutout, resizing, flipping and color jittering were all applied following the recent works \cite{peng2022few, tao2020few, zhang2021few}. Additionally, we adopted the standard data pre-processing as in \cite{peng2022few}. With a variation of number of epochs and iterations for base and incremental session across three datasets we keep the batch size of 512 for base session and 64 for incremental session during simplex finetuning as explained in Section \ref{csf}. Dataset wise epochs/iterations and learning rates for base and incremental session is given in Table \ref{tab:epc}. Additionally, we use SGDR \cite{loshchilov2016sgdr} with momentum as optimizer which uses cosine annealing strategy to reduce learning rate. Our code will be publicly available upon acceptance.
\begin{table*}[htb]
\centering
\scriptsize
\begin{tabular}{llllllllllllll}
\toprule
\multicolumn{1}{c}{\multirow{2}{*}{\textbf{Methods}}} & \multicolumn{11}{c}{\textbf{Session Accuracy (\%) ($\uparrow$)}}                                       & \multicolumn{1}{c}{\multirow{2}{*}{\textbf{\begin{tabular}[c]{@{}c@{}}Average \\ Acc. ($\uparrow$)\end{tabular}}}} & \multicolumn{1}{c}{\multirow{2}{*}{\textbf{\begin{tabular}[c]{@{}c@{}}Relative \\ Improvement\end{tabular}}}} \\ \cmidrule{2-12}
\multicolumn{1}{c}{}                                  & \multicolumn{1}{c}{\textbf{0}} & \multicolumn{1}{c}{\textbf{1}} & \multicolumn{1}{c}{\textbf{2}} & \multicolumn{1}{c}{\textbf{3}} & \multicolumn{1}{c}{\textbf{4}} & \multicolumn{1}{c}{\textbf{5}} & \multicolumn{1}{c}{\textbf{6}} & \multicolumn{1}{c}{\textbf{7}} & \multicolumn{1}{c}{\textbf{8}} & \multicolumn{1}{c}{\textbf{9}} & \multicolumn{1}{c}{\textbf{10}} & \multicolumn{1}{c}{}                                                                                                            & \multicolumn{1}{c}{}                                                                                          \\ \midrule
iCaRL \cite{rebuffi2017icarl}                                               &      68.68      &    52.65        &      48.61       &     44.16       &     36.62       &    29.52         &     27.83        &     26.26        &    24.01         &     23.89 &    21.16 &  36.67                                                                                                                          &    \textbf{+39.08}                                                                                                           \\
EEIL \cite{castro2018end}                                               &      68.68      &    53.63        &     47.91       &     44.20       &    36.30       &  27.46        &    25.93         &    24.70        &   23.95         &       24.13  & 22.11  & 36.27 &     \textbf{+38.13}                                                                                        \\

NCM \cite{hou2019learning}                                                 &     68.68      &    57.12         &        44.21     &    28.78        &    26.71         &   25.66        &     24.62       &     21.52       &    20.12       &     20.06  & 19.87 &     32.49                                                                                                                    &     \textbf{+40.37}                                                                                                          \\
Fixed classifier \cite{9413299}                                               &       68.47      &   51.00       &     45.42       &     40.76      &     35.90      &   33.18         &    27.23        &    24.24        &   21.18         &       17.34 &    16.20 &     34.63                                                                                                                &    \textbf{+44.04}                                                                                                           \\
 
D-NegCosine \cite{liu2020negative}                                               &       74.96     &    70.57        &     66.62       &     61.32       &     60.09       &   56.06         &    55.03       &    52.78       &  51.50         &       50.08   & 48.47 &   58.86                                                                                                                    &  \textbf{+11.77}                                                                                                             \\

D-DeepEMD \cite{zhang2020deepemd}  &     75.35     &     70.69         &     66.68       &     62.34       &     59.76       &   56.54         &    54.61        &    52.52       &   50.73         &  49.20   &  47.60 &   58.73                                                                                                                      &  \textbf{+12.64}                                                                                                             \\
D-Cosine \cite{vinyals2016matching}                                             &     75.52          &      70.95       &     66.46        &    61.20        &   60.86        &     56.88       &    55.40        &   53.49         &     51.94      &       50.93 &  49.31 & 59.36                                                                                                                        &    \textbf{+10.93}                                                                                                           \\
DeepInv \cite{yin2020dreaming}                                             &    75.90          &     70.21       &     65.36        &    60.14        &   58.79         &    55.88      &  53.21       &  51.27         &      49.38     &       47.11 & 45.67 &  57.54                                                                                                                        &    \textbf{+14.57}                                                                                                           \\ \midrule

TOPIC \cite{tao2020few}                                                &   68.68          &    62.49        &   54.81        &    49.99         &     45.25       &    41.40         &   38.35        &   35.36        &    32.22         &      28.31 &  26.28 & 43.92 &    \textbf{+33.96}                                                                                                           \\
IDLVQ \cite{chen2020incremental}                                                &    77.37         &    74.72       &       70.28      &     67.13      &    65.34       &  63.52         & 62.10         &    61.54        &      59.04      &    58.68  & 57.81 & 65.23                                                                                                                             &     \textbf{+2.43}                                                 \\
SPPR \cite{zhu2021self}                                                &    68.68         &    61.85       &       57.43      &     52.68      &   50.19        &  46.88        &  44.65         &    43.07        &      40.17      &    39.63  & 37.33 &  49.32                                                                                                                         & \textbf{+22.91}                                                                                                               \\
\cite{cheraghian2021synthesized}  &   68.78        &     59.37        &       59.32     &    54.96      &    52.58       &   49.81         &   48.09         &    46.32         &      44.33      &    43.43 &  43.23 & 51.84                                                                                                                               &   \textbf{+17.01}                                                                                                            \\

CEC \cite{zhang2021few}  &    75.85        &    71.94      &         68.50  &   63.50         &     62.43       &  58.27         &     57.73      &  55.81        &    54.83       &   53.52   & 52.28 &  61.33                                                                                                                          &    \textbf{+7.96}                                                                                                           \\
LIMIT \cite{zhou2022few}                                                  &  76.32          &   74.18         &   \textbf{72.68}         &   69.19         &    68.79        &   65.64        &    63.57       &   62.69         &   61.47         &     60.44 & 58.45 &  66.67                                                                                                                          &      \textbf{+1.79}                                                                                                         \\
MgSvF \cite{zhao2021mgsvf}                                                &   72.29          &   70.53         &   67.00         &    64.92      &    62.67       &     61.89    &     59.63        &    59.15         &   57.73          &       55.92   & 54.33 &  62.37                                                                                                                     &   \textbf{+5.91}                                                                                                            \\
MetaFSCIL \cite{chi2022metafscil}                                          &   75.9          &    72.41        &     68.78      &     64.78        &    62.96      &     59.99       &   58.3       &   56.85         &     54.78     &    53.82  & 52.64 &   61.93                                                                                                                        & \textbf{+7.6}                                                                                                              \\

FACT \cite{zhou2022forward}                                            &    75.90        &    73.23        &     70.84      &   66.13         &    65.56       &     62.15      &    61.74         &       59.83     &     58.41      &     57.89 & 56.94 & 64.42                                                                                                                         & 
 \textbf{+3.3}                                                                                                              \\

Data-free replay\cite{liu2022few}                                              &   75.90          &     72.14        &     68.64       &     63.76       &   62.58         &    59.11        &    57.82        &    55.89      &    54.92  & 53.58     &     52.39  &  61.52                                                                                                                             &   \textbf{+7.85}                                                                                                            \\
ALICE \cite{peng2022few}                                                &    77.40         &     72.70      &   70.60          &     67.20        &   65.90        &   63.40          &    62.90      &    61.90       &  60.50        &     \textbf{60.60} &  60.10 & 65.75                                                                                                                        &    \textbf{+0.14}                                                                                                           \\
SSFE-Net \cite{pan2023ssfe}                                             &      76.38       &     72.11        &     68.82         &      64.77       &     63.59        &    60.56         &      59.84      &    58.93        &    57.33          &    56.23 &  54.28      &  62.98                                                                                                                       &    \textbf{+5.96 }                                                                                                     \\

NC-FSCIL \cite{yang2023neural}                                             &      80.45        &     75.98       &     72.30        &     70.28        &     68.17        &   65.16         &  64.43          &      63.25      &    60.66      &     60.01 & 59.44 & 67.28                                                                                                                            &   \textbf{+0.8}                                                                                                            \\ \midrule
\textit{\textbf{MASIL(Ours)}}                         &    \textbf{80.50}        &      \textbf{76.02}      &    72.25       &      \textbf{70.30}      &    \textbf{68.85}        &    \textbf{65.72}        &    \textbf{64.45}        &    \textbf{63.28}        &    \textbf{60.80}        &    \textbf{60.60}   &  \textbf{60.24}  &            \textbf{67.54}                                                                                                            &                                                                                                               \\ \bottomrule
\end{tabular}
\caption{Performance comparison on CUB-200 with ResNet-18 as backbone architecture under \textit{10-way 5-shot} FSCIL setting. Table denotes the accuracy in each session, average accuracy across sessions and "Relative Improvement" denotes the improvement of our method in the last session. Methods above separating line are CIL methods for FSCIL as in \cite{tao2020few} and \cite{zhang2021few}}
\label{tab:3}
\end{table*}

\section{Additional Results}
\label{addres}
We continued summarising results for CUB-200 comparing various methods with \textbf{\textit{MASIL}}. Although improvement in  last session is very small +0.14\% as compared to strongest baseline ALICE \cite{peng2022few}. But we are consistently better in average accuracy and session wise accuracy (except only two sessions 2 and 8). On average accuracy we outperform ALICE \cite{peng2022few} by +1.79\% as shown in the Table \ref{tab:3}. 
To further analyze the underlying reason for performance improvement because of fine tuning of classifier weights obtained from concept basis, we calculated the average cosine similarity of the concept basis $c_i$ with all $c_j$, where $j \neq i$ for all the three datasets as given in the Table \ref{tab:tab4}. Formally it is calculated as:
\begin{equation}
\frac{1}{K(K-1)}\sum_{i=1}^{K}\sum_{j=1, j \neq i}^{K} c_i \cdot c_j
\end{equation}
\begin{table}[htb]
\centering
\begin{tabular}{l|c}
\toprule
\textbf{Dataset} & \textbf{Average Cosine Similarity} \\ \midrule
miniImageNet     &         -5.22e-4                                               \\
CIFAR-100        &         -8.78e-3                                              \\
CUB-200          &         -4.54e-4                                             \\ \bottomrule
\end{tabular}
\caption{Calculated cosine similarity among concept basis for each of the three benchmark datasets}
\label{tab:tab4}
\end{table}
\\
These entries are almost close to zeros resulting in the concept basis which are non-overlapping and non-repetitive and hence can induce the unique classifier weights correspond to novel classes, that can be represented in terms of their combination. 

\end{document}